\documentclass[journal]{IEEEtran}

\usepackage{times}
\usepackage{epsfig}
\usepackage{graphicx}
\usepackage{amsmath}
\usepackage{amssymb}
\usepackage{epstopdf}
\usepackage{xcolor}

\newcommand{\etal}{\textit{et al}.\:}
\newcommand{\ie}{\textit{i}.\textit{e}.\:}
\newcommand{\eg}{\textit{e}.\textit{g}.\:}

\newcommand{\norm}[1]{\left\lVert#1\right\rVert_2}

\ifCLASSINFOpdf
\else
\fi

\begin{document}
\title{Semantic Granularity Metric Learning for Visual Search}

\author{Dipu Manandhar$^{\star}$  \qquad Muhammet Bastan$^{\dagger}$   \qquad Kim-Hui Yap$^{\star}
  \thanks  {DM is currently with University of Surrey.}
  \thanks {This work was carried out while DM and MB were at Nanyang Technological University.}
   \IEEEauthorblockA{$\\$^{\star}$School of Electrical and Electronic Engineering, Nanyang Technological University, Singapore\\
  $^{\dagger}$Amazon, Palo Alto, California, USA \\ 
     \small{ Email: dipu002@e.ntu.edu.sg, mbastan@amazon.com, ekhyap@ntu.edu.sg}}}

\maketitle

\begin{abstract}
Deep metric learning applied to various multimedia applications has shown promising results in tasks such identification, retrieval and recognition. 
Existing metric learning methods often do not consider different granularity in visual similarity. However, in many domain applications, images exhibit  similarity at multiple granularities with visual semantic concepts, \eg fashion demonstrates similarity ranging from clothing of the exact same instance to similar looks/design or a common category.   
Therefore,  training image triplets/pairs used for metric learning inherently possess  different degree of information. 
However, the existing methods often treats them  with equal importance during  training. 
This hinders  capturing the underlying granularities in feature similarity which is required for effective visual search. 

In view of this, we propose a new deep semantic granularity  metric learning (SGML) that develops a novel idea of detecting and leveraging attribute semantic space to capture different granularity of similarity, and then integrate this information into deep metric learning.  
The proposed framework simultaneously learns image attributes and embeddings using multitask CNNs with shared parameters. The two tasks are not only jointly optimized but are further linked by the semantic granularity similarity mappings to leverage the correlations between the tasks. 
To this end, we propose a new soft-binomial deviance loss that effectively integrates the degree of information in training samples, which helps to capture visual similarity at multiple granularities. 
Compared to recent ensemble-based methods, our framework is conceptually elegant, computationally simple and provides better performance. 
We perform extensive experiments on benchmark metric learning datasets and demonstrate that our method outperforms recent state-of-the-art methods, \eg, 1-4.5\% improvement in Recall@1 over the previous state-of-the-arts \cite{kim2018attention,Cakir_2019_CVPR} on DeepFashion In-Shop dataset.  
\end{abstract}

\IEEEpeerreviewmaketitle

\section{Introduction}
\label{sec:intro}
\IEEEPARstart{D}{eep} learning and convolutional neural networks (CNNs) have achieved seminal results in multimedia applications such as image classification \cite{krizhevsky2012imagenet,he2016deep,horiguchi2018personalized}, object detection \cite{girshick2014rich,tang2019salient}, image
segmentation \cite{long2015fully,wang2019learning}, retrieval \cite{wan2014deep,DBLP:journals/corr/RazavianASC14,babenko2014neural,li2015weakly, bai2019deep}, cross-modal matching \cite{liong2016deep,kang2015learning,zhang2016cross} etc.
With this success, deep metric learning has recently been applied to  various recognition and visual search \cite{schroff2015facenet,shankar2017deep,bell2015learning,kiapour2015buy,li2015weakly,wu2018and}, one of the fundamental tasks in multimedia analysis.    
Deep metric learning aims to learn feature embeddings directly from raw image data together with associated distance metric.
Typically, CNNs are used to learn the non-linear transformations that preserve  visual similarity in the embedding space.  
The key idea is to map images from the same class having similar semantic meaning into closer points in embedding space, while that from different classes are separated farther apart.

\begin{figure}[t!]
\centering
\includegraphics[scale=0.25,  trim={10cm 7cm 3.3cm 5.5cm},clip]{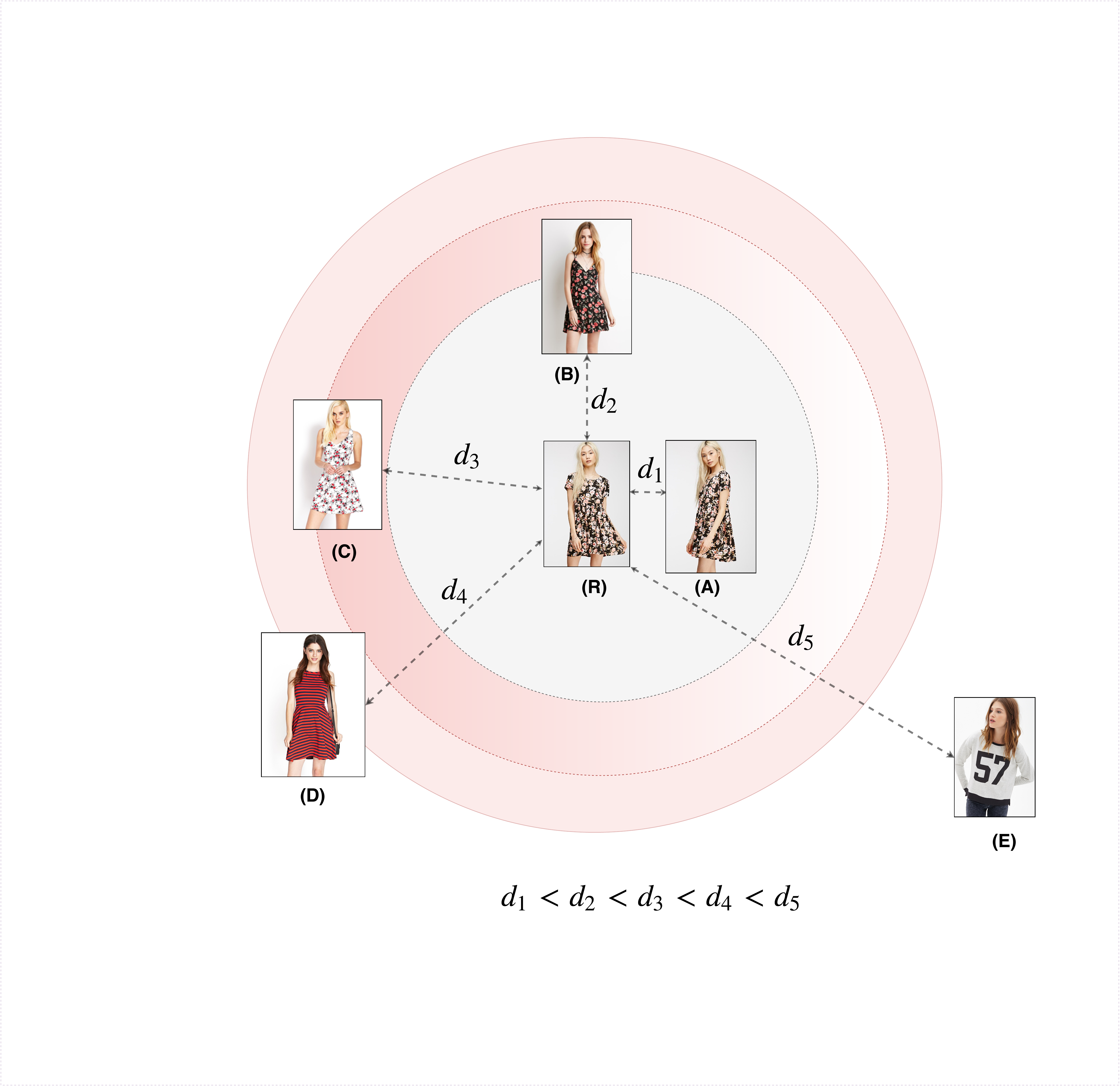}
\caption{Illustration of desirable feature space for visual search. Images exhibit similarities at multiple levels such as exact same instance, same category, or similar visual looks based on semantic concepts related to attributes such as \emph{pattern, color, sleeve length, clothing category}.}
\label{fig:feature_space}
\end{figure}

Fig.~\ref{fig:feature_space} illustrates the feature space desirable for good retrieval performance. It can be observed that image similarity can be defined at multiple levels depending upon the granularity of semantic similarity. 
For example, with respect to the reference image (R), image (A) is the same instance and shares the same semantic attribute space \eg \emph{black, floral, half-sleeve, dress}. They have the highest level of similarity and lie closest to each other. The second closest image (B) is semantically similar and have many common attributes \emph{black, floral dress} with minor differences in design. Subsequently, image (C)  represent a \emph{floral dress} with different color, and image (D) represent different image from dress category, and hence they lie farther apart in the feature space. Finally, out-of-category image (E) lie the farthest.  Therefore, semantic concepts define the visual similarity at different granularities  which should be effectively captured in feature space for a good retrieval performance.       

\begin{figure}[t!]
\centering
\includegraphics[scale = 0.38]{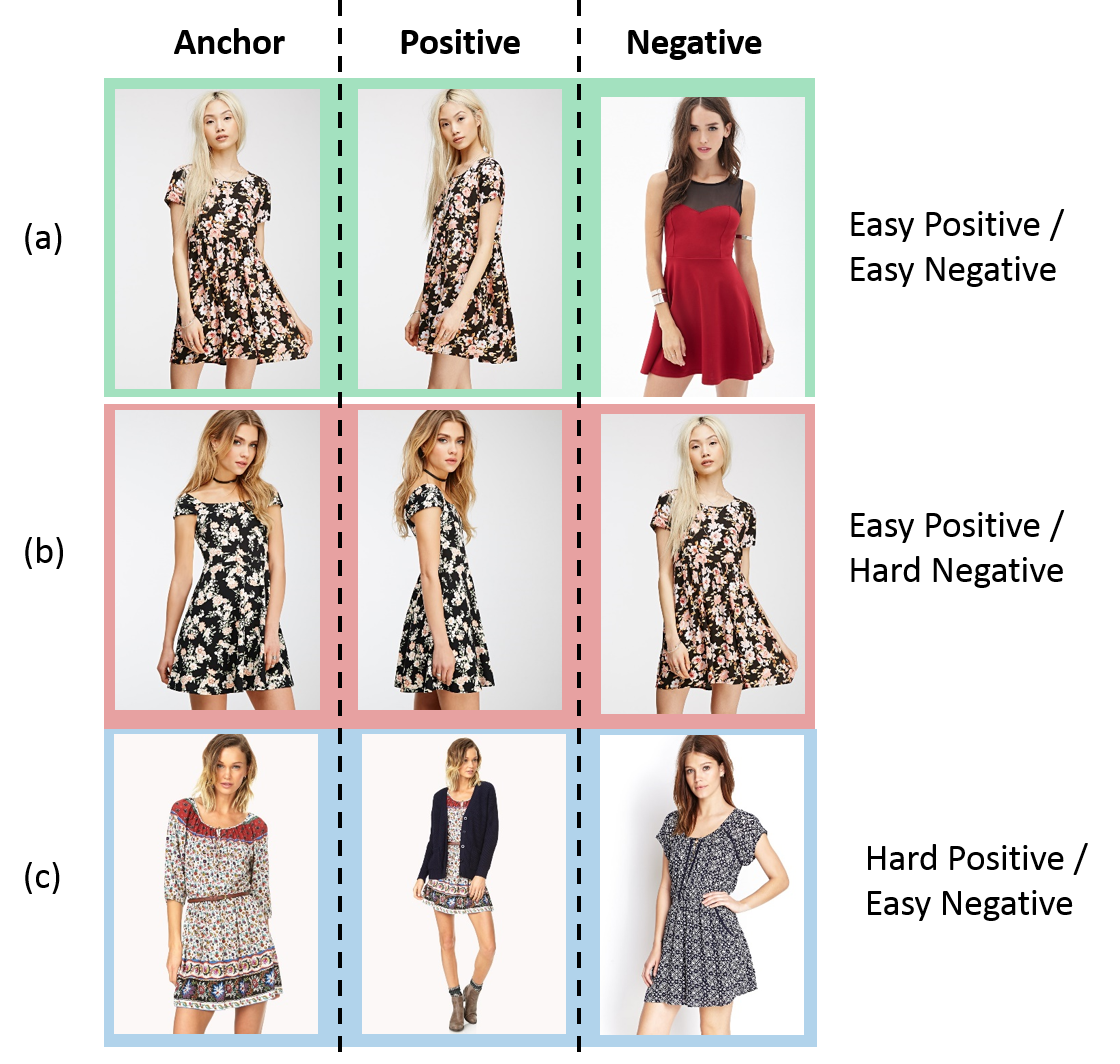}
\caption{Various types of training pairs/triplets from DeepFashion-Inshop dataset \cite{liu2016deepfashion} with semantic similarity at different granularities. Each row forms two pairs or a triplet, and  have different levels of information. Easy examples are less informative whereas hard examples are more informative.}
\label{fig:triplet_eq}
\end{figure}

Generally, metric learning networks such as Siamese networks \cite{bromley1994signature} and triplet networks \cite{schroff2015facenet} are trained based on pairs \cite{bell2015learning, bromley1994signature} and triplets \cite{schroff2015facenet,hermans2017defense} respectively to learn the feature similarity space. 
However,  optimizing a metric learning is  difficult compared to other conventional tasks (e.g. classification), as it requires sampling image triplets/pairs for training. 
This poses huge challenges for metric learning especially due to two reasons. First, possible combinations for pairs/triplets grow quadratically/cubically  with respect to the number of images in the dataset. Second but most importantly, all training samples are not equally informative, rather a large portion of the samples are highly redundant which do not contribute to learning. 
For example, Fig.~\ref{fig:triplet_eq} shows example training samples from DeepFashion dataset \cite{liu2016deepfashion} where each row can be considered as a triplet or two pairs: one positive and one negative. 
The first row Fig.~\ref{fig:triplet_eq}(a) shows easy examples where images from the same class (positive) and from a different class (negative) have clear and distinct semantic differences, and hence easy to differentiate. Large portion of training samples falls under this category. During  training,  network quickly learns to differentiate them, and most of such samples become uninformative for further learning. 
The second row Fig.~\ref{fig:triplet_eq}(b) presents an easy positive but a hard negative, where the negative image from a different class shares high semantic similarity with the anchor image. Such samples are more informative and hence identifying them and learning the subtle differences is desirable for good retrieval accuracy. 
Likewise, the third row Fig.~\ref{fig:triplet_eq}(c) shows an informative hard positive example, where a portion of the (positive) clothing is occluded. Therefore, training samples inherently possess different degree of information depending upon the granularity of semantic similarity between images. 

In order to mine good training examples, the work in \cite{hermans2017defense,schroff2015facenet,wu2017sampling} used informative sample selection strategies. 
However, all these methods employ a hard binary decision during the selection \ie either to use the samples for training or completely discard them.  
Most importantly, all the selected samples are treated with equal importance to compute the metric loss. 
This creates restrictions to learn visual similarity at multiple granularities required for good retrieval performance as shown by the feature space in Fig.~\ref{fig:feature_space}. Other metric learning methods such as  \cite{huang2015cross, liu2016deepfashion, khamis2014joint} used image attributes as standard multitasks, however, they do not consider the possible interactions between attribute and metric learning. Hence, these methods do not effectively capture the underlying semantic granularity in image similarity.    

In view of this, we propose a new deep \emph{Semantic Granularity Metric Learning (SGML)} 
framework that detects and captures the attribute space semantic context in images  and then effectively incorporate this information into the metric learning. The key idea is to simultaneously estimate the degree of information in training samples during the metric learning to capture the similarity at multiple granularities and boost the retrieval performance.   
We summarize the main contributions of this paper as follows. 
We propose a new SGML network with a fusion multitask and metric learning which leverages interaction between semantic attribute learning and metric learning to capture underlying granularities in visual similarity. 
We measure the informativeness of the training samples using \emph{Semantic Granularity Similarity (SGS)} mapping and automatically identify  the informative samples during the training. Further, we propose a new
soft-binomial deviance loss which is dynamically modulated in a soft-manner based on the semantic granularity similarity.
Compared to recent ensemble and attention based metric learning methods, the SGML framework is elegant while providing better retrieval performance. 
Finally, our proposed SGML method sets new state-of-the-art retrieval results on the benchmark DeepFashion-Inshop dataset \cite{liu2016deepfashion} and CUB-200-2011 birds dataset \cite{WelinderEtal2010}.

\section{Related Work}
\label{sec:related_works}

\subsection{Deep Metric Learning}
Deep metric learning aims to learn a function $f(\cdot)$: $\mathbb{R}^N \rightarrow \mathbb{R}^d$ that maps images into an embedding space where visually similarity between images can be directly compared using distance between the image embeddings. 
In order to learn such embedding space, CNNs are trained using loss functions based on pairs/triplets of images.
One of the earliest works was carried by Bromley \etal \cite{bromley1994signature} which used deep Siamese networks with the contrastive loss for signature verification, which was later adapted for face verification \cite{chopra2005learning} and dimensionality reduction \cite{hadsell2006dimensionality}. Such contrastive loss \cite{bromley1994signature,chopra2005learning} poses a constraint that requires positive pairs to be mapped into the same point, and negative pairs to be separated at least by a fixed margin. Recently, the contrastive loss has been used for learning product similarity \cite{bell2015learning}, matching user photo with online products \cite{wang2016matching}, and face verification \cite{hu2014discriminative}. Different from this, triplet loss \cite{schroff2015facenet, hermans2017defense} poses a relative constraint \cite{chechik2010large} which requires the distance between the negative pairs to be greater than that between positive pair at least by a margin. 
Triplet loss has been used for face recognition and verification \cite{schroff2015facenet},  person re-identification \cite{ahmed2015improved,wang2016joint}, fine-grained classification \cite{qian2015fine}, feature learning \cite{lai2015simultaneous}, image ranking \cite{wang2014learning} and fashion search \cite{liu2016deepfashion,kiapour2015buy,shankar2017deep}.  While contrastive  and triplet losses have non-continuous gradient, binomial deviance loss \cite{esl-book} is a smooth function and provides continuous gradients, and hence it  has shown to achieve superior performance than other metric losses in recent studies \cite{opitz2017bier,bier-pami18}. More recently, FastAP \cite{Cakir_2019_CVPR} proposed ranking-based formulation that optimized average-precision of ranked lists for metric learning.

\subsection{Informative Training Sample Selection}
In order to effectively train metric learning networks, several works \cite{simo2015discriminative, schroff2015facenet, song2016deep, balntas2016learning,hermans2017defense, ge2018deep} have used  techniques for mining informative training samples. 
Hard negative mining \cite{simo2015discriminative} used the hardest negatives from a randomly selected set that gives the highest loss. 
Similarly,  FaceNet  \cite{schroff2015facenet} proposed semi-hard negative mining which is widely adopted in \cite{song2016deep,parkhi2015deep}. 
Song \etal \cite{song2016deep} proposed to lift features in a mini-batch to compute distance matrix and take full advantage of all possible pairs/triplets.
In \cite{balntas2016learning}, the authors proposed in-triplet negative mining where the positives and anchors are swapped to ensure the largest possible loss for the  given triplet.  
Hermans \etal \cite{hermans2017defense} proposed batch-hard training where all the triplets in a mini-batch are mined and only hard examples are used to compute the loss.
The authors in \cite{wu2017sampling} proposed distance-weighted sampling where negative samples are selected uniformly according to their distances. 
Similarly, the work in  \cite{huanglearning,tiered_ECCVW2018} used attribute information to sample the informative triplets and  weight the triplet loss.
Recently, Ge \etal \cite{ge2018deep} proposed anchor-neighbor sampling where a class distance matrix is computed at every epoch, and images from the nearest classes are used as negative samples for the subsequent epoch. However, it is computationally expensive to compute large distance matrix every epoch. Most importantly, all the selected samples having different level of information are treated with equal importance which hinders in capturing the desired similarity space as shown in Fig~\ref{fig:feature_space}.

\subsection{Metric Learning with Multitask Training}
Rather than training with a metric loss alone, several works \cite{yi2014learning,sun2014deep,schroff2015facenet,huang2015cross,liu2016deepfashion} have used multitask networks to perform joint optimization of metric learning and auxiliary tasks. 
A combination of contrastive loss and softmax loss have been used for face identification-verification in \cite{yi2014learning,sun2014deep}. 
Triplet loss and softmax loss  have been used for face recognition \cite{schroff2015facenet}  and  fine-grained object recognition \cite{zhang2016embedding}.   
Khamis \etal in \cite{khamis2014joint} used attribute learning and triplet loss  for person re-identification. 
Similarly, Siamese networks  with the integration with clothing attributes have been used in \cite{huang2015cross} for street-to-shop clothing matching. 
Recently, Liu \etal  \cite{liu2016deepfashion} proposed to jointly optimize classification, attribute detection and the triplet loss for visual fashion search. 
However, these methods perform standard multitask training but lack interactions among the tasks. As opposed to this, we further interlink the tasks, where semantic attribute space similarity is first used to automatically identify informative training samples and then dynamically learn the proposed soft-binomial deviance loss, leading to notable improvements in retrieval performance.

\subsection{Ensemble-based Deep Metric Learning}
Recently, several researchers in \cite{yuan2017hard, opitz2017bier, kim2018attention,xuan2018deep} have used ensemble techniques in metric learning to boost the retrieval accuracy.
Yuan \etal \cite{yuan2017hard} used models with different complexities obtained by networks of different depths and ensembled them  in a cascade form. 
Opitz \etal \cite{opitz2017bier} divided final embedding into non-overlapping representations and trained them using online gradient boosting to obtained the ensemble model. 
The work in \cite{kim2018attention} used attention-based ensemble where different learners attend to different parts of images. Similarly, the recent work \cite{xuan2018deep} used random bagging of training classes into meta-classes and trained multiple CNNs. The final embedding is then obtained by concatenating the embedding from individual networks. It used as much as 48 ResNet18 CNN models to form the ensemble. Hence, it can be quite expensive in terms of both memory and computation. Compared to these ensemble based methods,  the proposed SGML uses a single model while providing higher retrieval performance. 

\section{Proposed Semantic Granularity Metric Learning (SGML) Framework}
\label{sec:proposed_method}

\begin{figure*}[!t]    
    \centerline{\includegraphics[width=0.7\textwidth]{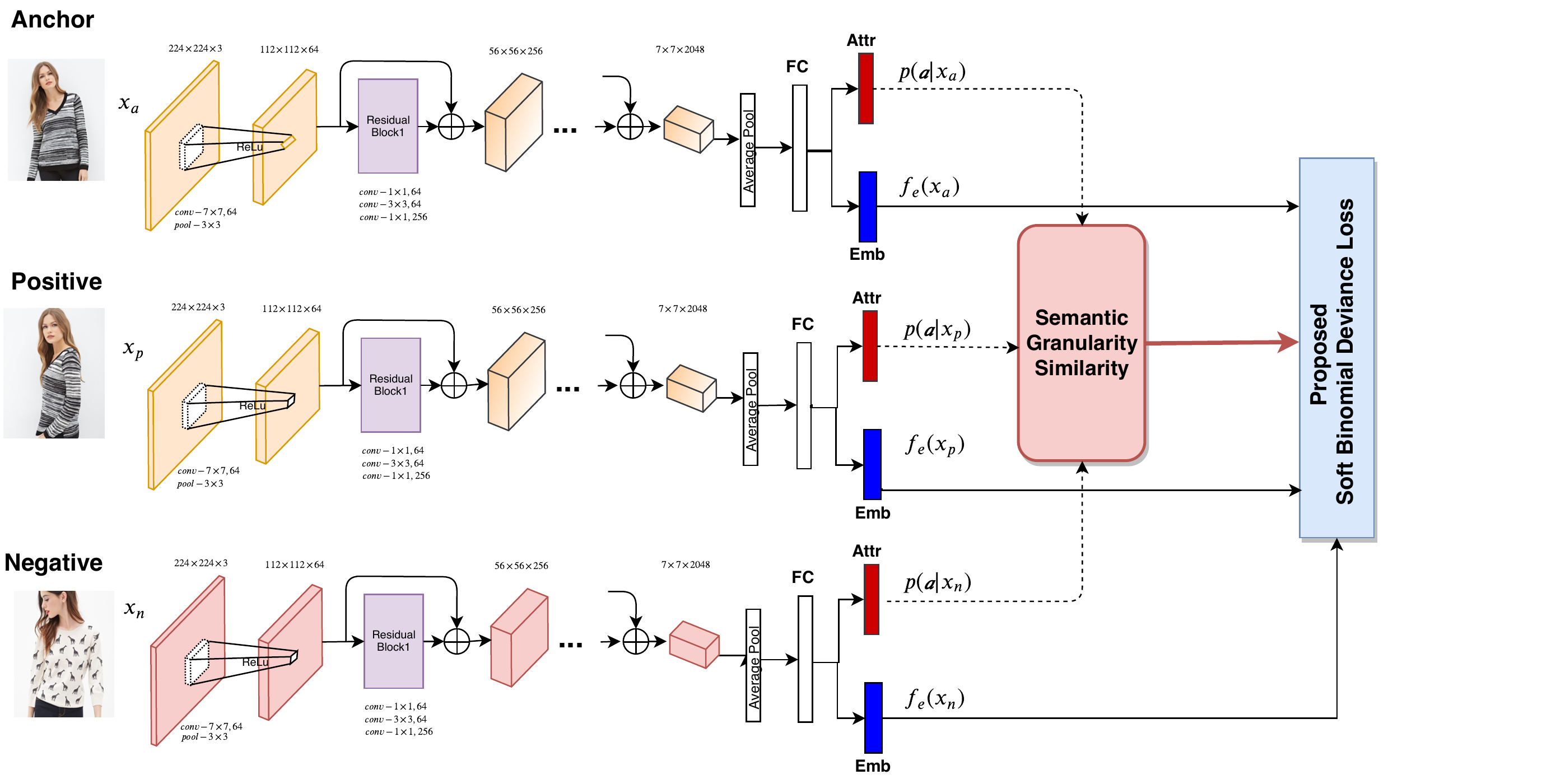}}    
    \caption{Overview of Semantic Granularity Metric Learning (SGML) framework. The shared CNN networks have fully connected layer with two branches for joint attribute and embedding learning. The proposed soft-binomial deviance loss is dynamically learned taking  semantic granularity similarity into account.}
    \label{fig:sgml}
\end{figure*}

\subsection{Overview of the Proposed SGML Framework}
Fig.~\ref{fig:sgml} shows an overview of the proposed semantic granularity metric learning (SGML) framework. The model consists of CNN networks with shared parameters for mapping  image triplets ($\boldsymbol{x_a}$, $\boldsymbol{x_p}$, $\boldsymbol{x_n}$) into feature embeddings.
The fully-connected (FC) layer has two branches: `Attr' branch for learning attributes $p(\boldsymbol{a}|\boldsymbol{x})$ and `Emb' branch for learning embedding  $f_{e}(\boldsymbol{x})$. 
Most interestingly, the two tasks are interlinked by the semantic granularity similarity (SGS) mapping of the positive and negative pairs obtained using simultaneously learned attribute information. During the training, the SGS mappings are first used to estimate the informativeness of training samples, and then to dynamically learn an appropriate metric loss using the proposed soft-binomial deviance function. This, in turn, helps to capture the visual similarity at multiple granularities  and boost the retrieval performance. Overall, the network is trained using $L_{\text{SGML}}(\theta)$ loss.
\begin{equation}
L_{\text{SGML}}(\theta) = L_{\text{SBDL}}(\theta) + \lambda L_{\text{BCE}}(\theta),
\end{equation}
where $\theta$ represents learnable parameters of the network, $L_{\text{SBDL}}(\theta)$  is the proposed soft-binomial deviance loss for metric learning,  $L_{\text{BCE}}(\theta)$ is binary cross-entropy loss in attribute learning, and  $\lambda$ is a factor that balances contribution of the two losses.  We describe these tasks and associated loss functions in the following sections.

\subsection{Attribute Prediction Task}
We first describe the attribute prediction task associated with the `Attr' branch.
We predict attributes present in the images such as \emph{color, texture, sleeve-length} etc. for clothing; and \emph{beak-shape, wings-pattern, tail-length} etc. for birds, which have been found instrumental in describing images and objects in a semantic context \cite{liu2016deepfashion, huang2015cross,lampert2014attribute,yang2016deep}. 

Formally, let $\boldsymbol{a} = [ a_1, a_2, \dots, a_K ]$ be  $K$ binary attributes of an image $\boldsymbol{x}$ where $a_i \in \left\lbrace 0,1 \right\rbrace$ denotes presence or absence of $i^{th}$ attribute.  
To predict the  attributes, the feature from the attribute branch (`Attr') is first passed through a sigmiod layer to squash the values into [0,1] which can be interpreted as the probabilities that each  attribute is present in the image \ie  $p(\boldsymbol{a}|\boldsymbol{x})$. We treat this problem as multi-label classfication and train the network using binary-cross entropy (BCE) loss.
\begin{align}
\label{eg:bce_loss}
	 L_{\text{BCE}}(\theta) = - \sum_{i=1}^K \left[a_i \log(p(a_i|\boldsymbol{x})) + (1-a_i) \log(1- p(a_i|\boldsymbol{x})) \right],
\end{align} 
where $a_i$ is ground-truth attribute label and  $p(a_i|\boldsymbol{x})$ is the corresponding predicted probability. Overall, the output  $p(\boldsymbol{a}|\boldsymbol{x})$ describes the appearance of image in attribute space. At this point, we can train a CNN using this BCE loss alone and perform retrieval using the FC layer feature from the network. However, such feature may not be optimal for retrieval tasks. 
In this work, we aim to learn discriminative embeddings with fusion of multitask and metric learning  where the predicted attributes are further exploited to capture semantic granularity in visual similarity.

\subsection{Semantic Granularity Driven Metric Learning}
\begin{figure}[t]    
    \centerline{\includegraphics[width=0.38\textwidth]{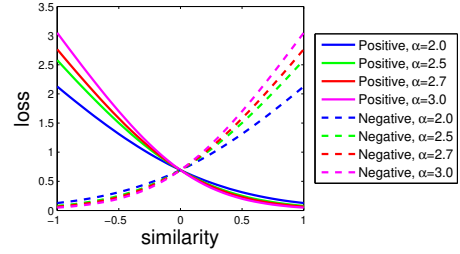}}    
    \caption{Binomial deviance loss function for positive and negative pairs  with different $\alpha$ values. As $\alpha$ increases, the curve gets steeper.}
    \label{fig:bindev-loss}
\end{figure}

In this section, we describe our proposed  approach in SGML framework  to learn  embedding $f_{e}(\boldsymbol{x})$ that captures  visual similarity at multiple granularities as shown in Fig.~\ref{fig:feature_space}. Different from existing methods \cite{schroff2015facenet,opitz2017bier,yi2014deep,bier-pami18,ge2018deep,hermans2017defense}, our  method integrates semantic granularity in visual similarity into metric learning.
We formulate our framework using binomial deviance-based loss~\cite{esl-book} that has a smooth function with continuous gradient. The original binomial deviance loss (BDL)  is given by 
\begin{equation}
L_{\text{BDL}}(s,y) = \log(1 + e^{-(2 y - 1) \alpha (s - \beta) C_y }),
\end{equation}  
where $y=1$ for positive pairs and $y=0$ for negative pairs; $\alpha$ and $\beta$ are scaling and translation parameters; $C_y$ is a cost factor to balance the positive and negative pairs.   
The loss function uses cosine similarity $s = s(f_{e}(\boldsymbol{x_1}), f_{e}(\boldsymbol{x_2}))$  between the embeddings of pairs of images $\boldsymbol{x_1}$, $\boldsymbol{x_2}$ as given by Eq.~(\ref{eq:cosine}). Fig.~\ref{fig:bindev-loss} shows the loss function for positive and negative pairs which  aims to maximize the similarity between positive pairs and minimize the similarity between negative pairs.  
\begin{align}
s(f_{e}(\boldsymbol{x_1}), f_{e}(\boldsymbol{x_2})) =  \frac{f_{e}(\boldsymbol{x_1})^{\top}f_{e}(\boldsymbol{x_2})}{\norm{f_{e}(\boldsymbol{x_1})}  \norm{f_{e}(\boldsymbol{x_2})}}.
\label{eq:cosine}
\end{align}
However, this metric loss function is entirely based on embedding similarity. Thus, all the training samples are treated in the same manner and given equal importance. Hence, we propose a new soft-binomial deviance loss that not only considers embeddings but also the predicted attributes into metric loss to capture the semantic granularity in visual similarity.  

Formally, let $( \boldsymbol{x_a}$, $\boldsymbol{x_p}$, $\boldsymbol{x_n})$ be an input triplet to the proposed network in Fig.~\ref{fig:sgml}, where $(\boldsymbol{x_a}$, $\boldsymbol{x_p})$ forms a positive pair ($y=1$); and  $(\boldsymbol{x_a}$, $\boldsymbol{x_n})$ forms a negative pair ($y=0$).    
Let the corresponding output embeddings from the branch `Emb' be $\left(f_e(\boldsymbol{x_a}), f_e(\boldsymbol{x_p}), f_e(\boldsymbol{x_n})\right)$. 
Similarly, the outputs of the attribute branch `Attr' be $\left(p(\boldsymbol{a}|\boldsymbol{x_a}), p(\boldsymbol{a}|\boldsymbol{x_p}), p(\boldsymbol{a}|\boldsymbol{x_n})\right)$ which are  attribute distributions that describe images in semantic attribute space.
We compute the pairwise similarities between the  embeddings using the cosine function in Eq.~(\ref{eq:cosine}). Let $s(a,p) = s (f_e(\boldsymbol{x_a}), f_e(\boldsymbol{x_p}))$ and $s(a,n) = s(f_e(\boldsymbol{x_a}), f_e( \boldsymbol{x_n}))$  be the similarities between the positive  and negative embeddings respectively. 
These cosine similarities between the embeddings lie in $[-1,+1]$, enforcing an upper bound on the loss magnitudes and facilitating the optimization process. 

We further define semantic granularity similarity (SGS) mapping of the pair of images using the predicted attribute distributions $p(\boldsymbol{a}|\boldsymbol{x})$. The SGS mapping $g(\boldsymbol{x_1}, \boldsymbol{x_2})$ between image $\boldsymbol{x_1}$ and $\boldsymbol{x_2}$ is given by
\begin{align}
g(\boldsymbol{x_1}, \boldsymbol{x_2}) = \frac{\sum_{i=1}^K p(a_i|\boldsymbol{x_1}) \, p(a_i|\boldsymbol{x_2})}{ \sqrt{\sum_i p(a_i|\boldsymbol{x_1})^2} \sqrt{\sum_i p(a_i|\boldsymbol{x_2})^2 } }, 
\end{align}
where $p(a_i|\boldsymbol{x})$ is the probability of attribute $i$ in image $\boldsymbol{x}$. The value of $g$ lies between 0 and 1, which indicates the degree of similarity  in semantic attribute space. Without loss of generality, let $g(a,p) = g(\boldsymbol{x_a}, \boldsymbol{x_p})$ and $g(a,n) = g(\boldsymbol{x_a}, \boldsymbol{x_n})$ be the SGS mappings of the positive and negative image pairs respectively.

Using above definations, we define our proposed soft binomial deviance loss (SBDL)  as follows.
\begin{align}
L_{\text{SBDL}} &= \frac{1}{M} \sum_i^M L^{(i)}_{\text{pos}} +  \frac{1}{N} \sum_j^N L_{\text{neg}}^{
(j)} 
\end{align}
where,
\begin{align}
L_{\text{pos}}(s,g |y=1) &= \log \left( 1 + e^{- \alpha [(s(a,p) + g(a,p) - \beta] }\right), \label{eq:pos_sbdl} \\
L_{\text{neg}}(s,g |y=0) &= \log \left(1 + e^{+\alpha [s(a,n) - g(a,n) - \beta] } \right). \label{eq:neg_sbdl}
\end{align}

The proposed soft-binomial deviance loss  $L_{\text{SBDL}}$ is the sum of the average losses over $M$ positive and $N$ negative image pairs, where $\alpha$ and $\beta$ are scaling and translation parameters.  Note that  the proposed loss functions in Eq.~(\ref{eq:pos_sbdl}) and (\ref{eq:neg_sbdl})  integrates semantic attribute information into the metric learning.
 
\begin{figure}[t]    
    \centerline{\includegraphics[width=0.30\textwidth]{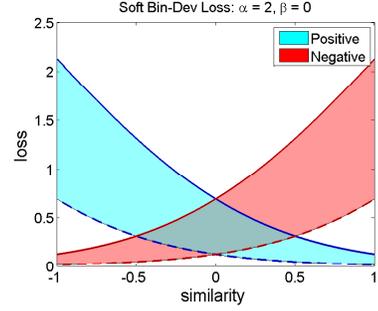}}    
    \caption{Proposed SBDL loss function for positive and negative pairs for embedding similarity $s$ in horizontal axis. The semantic granularity similarity (SGS) defines shaded regions where the  loss is dynamically modulated.}
    \label{fig:soft-bindev-loss}
\end{figure}

Fig.~\ref{fig:soft-bindev-loss}  shows the proposed SBDL loss functions where the horizontal axis represents the embedding similarity. 
The loss maximizes the embedding similarity ($s$) between the positive pairs and minimizes that between the negative pairs.
Different from the single line binomial deviance loss curves in Fig~\ref{fig:bindev-loss}, the proposed SBDL loss function forms bands which 
allows to dynamically modulate the metric loss within the shaded region.  
For both positive pairs (shaded blue) and negative pairs (shaded red), the SBDL adaptively learns the metric loss based on garnularity of semantic similarity, where the minimum value of the SGS mapping sets the upper limits of the bands and vice-versa.  More precisely, for the positive pairs, the larger the positive SGS mapping $g(a,p)$, the lower is the loss. For example, the positive pair in Fig.~\ref{fig:triplet_eq}(a) forms an easy positive example which have large semantic similarity, and hence a lower loss is imposed for such easy and confident positive pairs. Inversely, for the hard positive examples (Fig.~\ref{fig:triplet_eq}(c)),  larger loss is imposed to pull them as close as possible.
On the other hand, when  the  negative SGS mapping $g(a,n)$ is larger, they instead form a hard negative pair. For instance, the negative pair in  Fig.~\ref{fig:triplet_eq}(b) forms a hard negative example as they share many attributes (\emph{floral, dress, short-sleeve} etc).  Although  they are from different classes, they are semantically similar, and hence  a lower loss is incurred for such negative pairs to keep them relatively closer in the feature space. Overall, the proposed SGML framework adaptively  modulates the metric loss and hence training samples are treated based on granularity of semantic visual similarity. This helps to capture the feature embedding space at multiple granularities and boost the retrieval performance.

\section{Sampling Methods in Deep Metric Learning}
\label{sec:samplingmethods}
Sampling strategies used to select training examples may impact the performance of metric networks. 
Typically, for an anchor image from a reference class, image pairs are sampled randomly:  the positive being sampled from the pool of images from the same class, and the negative from the pool of different classes. In class retrieval problem, each class generally contains sufficient number of images for sampling (\eg 50-60 images per class in CUB \cite{WelinderEtal2010} dataset).
However, in many practical applications such as clothing instance retrieval in DeepFashion \cite{liu2016deepfashion}, there are large number of classes and only one or a few relevant images in each class. This leads to a severe unbalance between the number of candidate positive and negative pairs. In this paper, we investigate two strategies to address the potential issues during the sampling: \emph{Image-wise Sampling} and \emph{Batch-wise Sampling}.   

\begin{figure}[!t]
\centering
\includegraphics[scale=0.5]{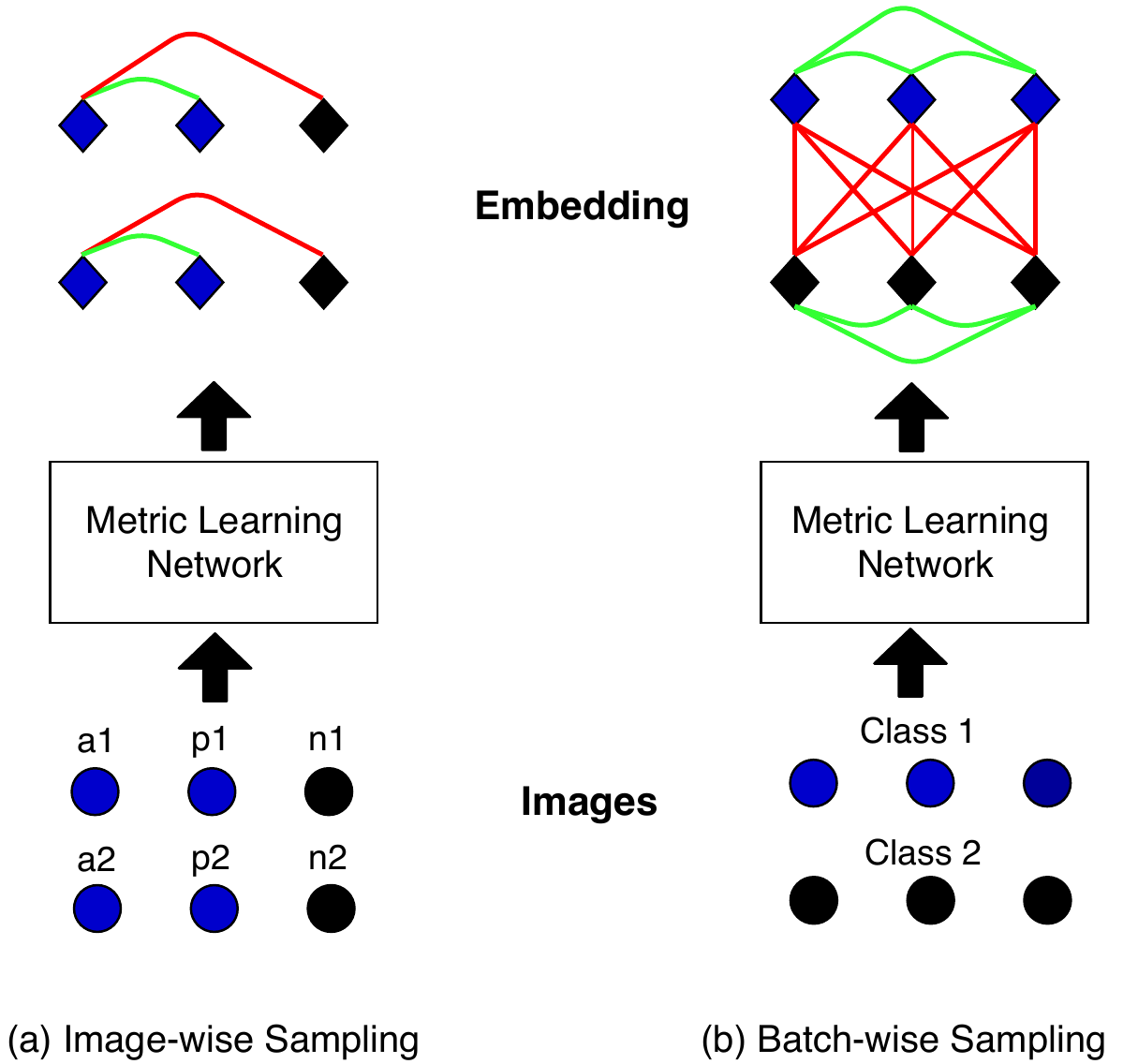}
\caption{Sampling strategies in metric learning. (a) Image-wise sampling pre-defines  positive and negative pairs on images; (b) Batch-wise sampling performs pairing on  embeddings, and all the possible pairs are considered. Same color represent images/embeddings from the same class; positive and negative pairs are represented by green and red connections respectively.}
\label{fig:sampling_methods}
\end{figure}

The image-wise sampling method uses each training image as anchor, then randomly select one positive image from the same class, and one negative image from a different class \cite{schroff2015facenet,wang2014learning}. Hence, each mini-batch has a total of 3$N$ images, $N$ being number of anchor images considered. Different from this, the batch-wise sampling first randomly select $N'$ classes and sample $M'$ images from each class which leads to $N'M'$ images in a mini-batch. The pairing then performed on the embeddings, and all possible positive and negative pairs within this batch are taken into account.
Fig.~\ref{fig:sampling_methods} illustrates the two types of sampling methods, where a total of 6 images is used for both methods. Image-wise sampling pre-defines the pairs on images and hence forms 2 positive and 2 negative pairs. The batch-wise sampling instead performs pairing on embeddings and considers all the possible pairs \ie 6 positive pairs and 9 negative pairs,  larger number of pairs compared to image-wise sampling. We conduct experiments using both strategies and provide inferences for  the choice of sampling methods to achieve good retrieval performance. 

\section{Experiments}
\label{sec:exp}
We performed extensive experiments on standard benchmark datasets, and report state-of-the-art results which  demonstrate the effectiveness of the proposed SGML framework. 
. 

\subsection{Datasets}
\label{ssec:datasets}
\subsubsection{DeepFashion-Inshop Dataset \cite{liu2016deepfashion}}
\label{sssec:df}
DeepFashion-Inshop dataset  contains fashion images with  clothing ID associated with each class. It has  25,882 images from 3997 classes  for training,  12,612 gallery and 14,218 query images from 3,985 classes.  Given a query image, the goal is to the retrieve gallery images with the same class ID as the query.
Each of the classes is annotated with 463 binary clothing attributes (\eg \emph{clothing fabric,  pattern, neck-shape}) which describes the fashion images. Following the standard practice, we use Recall@K as the performance metric where $K \in \left\lbrace 1,10,20,30,40,50\right\rbrace$.

\subsubsection{CUB-200-2011 Dataset \cite{WelinderEtal2010}}
CUB-200-2011 (CUB) dataset  contains 11,788 images of 200 birds species. Each image is annotated with class labels and 312 binary attributes (\eg \emph{belly color, beak shape}). Following common practice, we use the first 100 classes (5,864 images) for training, and remaining 100 classes (5,924 images) for testing. Given a query image, the goal is to retrieve images of the same species from the remaining images of the test set. This is a class retrieval problem. 
Following the standard protocol, we crop the images using the provided bounding boxes to remove background clutters, and measure the retrieval performance using Recall@K, where $K \in \left\lbrace1,2,4,8,16,32\right\rbrace$.

\subsection{Experimental Settings}
\label{ssec:exp}
We used the proposed SGML architecture as shown in Fig.~\ref{fig:sgml} with ResNet101 \cite{he2016deep} as CNN backbone which has less parameters and provides faster convergence compared to other CNNs such as VGGNet \cite{simonyan2014very}.   
The output of the CNN backbone of size 2048 is fed to a fully-connected (FC) layer of size 1024 which has  two branches: one for embedding learning with dimension 512 and the other for attribute learning with dimension equal to the number of attributes in the dataset. We performed data augmentation similar to  \cite{song2016deep,bier-pami18}. We first cropped the images using random scaling (0.9 to 1.0) of original image, which are then resized  to $224 \times 224$ together with random horizontal flips. During the training, online  sampling is performed such that images from the class form  positive pairs and that from different classes form negatives pairs. We used a mini-batch size of $3 \times 60$ for image-wise sampling and a mini-batch size of 164 $(N' = 41$ and $M' =4)$ for batch-wise sampling.
The network is trained using Adam optimizer with initial learning of  $10^{-4}$  and $4\times 10^{-6}$ for  DeepFashion dataset and CUB dataset respectively. A lower initial learning rate is used for CUB dataset to prevent overfitting during the training, as the dataset is relatively small.
We explored a range of values for $\alpha$ and $\beta$ for soft-binomial deviance loss, and used $\alpha =2$ and $\beta = 0.5$ for DeepFashion dataset; and $\alpha=3$ and $\beta=0.1$ for CUB dataset in all experiments unless specified. We also performed thorough experiments to study the impact of these parameters on the retrieval performance. The algorithm is implemented using PyTorch framework \cite{paszke2017automatic}. 

In the following sections, we show how our proposed SGML framework improves the  performance and outperforms the state-of-the-art methods. We also present analysis on impact of various factors such as  choice of sampling methods and embedding layer, parameters of the proposed SBDL loss including ablation studies that demonstrate the superiority of SGML when compared to strong baselines. 

\subsection{Results: DeepFashion Dataset}
\subsubsection{Comparison with State-of-the-art Methods}
We compare the proposed SGML with several methods using various loss functions \cite{kiapour2015buy,opitz2017bier,ge2018deep}, multitask methods \cite{liu2016deepfashion, huang2015cross}, and recent ensemble-based methods \cite{kim2018attention, bier-pami18,yuan2017hard,xuan2018deep}	 that use moderate to large number of models.  Table~\ref{table:df_sota} quantifies the performances of the existing state-of-the-arts and the proposed SGML framework. 
SGML clearly outperforms methods using popularly used constrastive loss \cite{kiapour2015buy}, triplet loss (HTL) \cite{ge2018deep}, and binomial deviance loss \cite{opitz2017bier} by large margins ($>$10\% Recall@1) which demonstrates the advantage of the proposed soft-binomial deviance loss (SBDL). 
Similarly, the proposed SGML outperforms  the exiting multitask approaches such as Dual Attribute-Aware Network (DARN) \cite{huang2015cross} and FashionNet \cite{liu2016deepfashion} that used joint attribute and metric learning. 
Compared to these methods, SGML allows interactions between the tasks and hence inherently considers different granularities of semantic similarity.
The top-performing methods  such as hard-aware cascaded network (HDC) \cite{yuan2017hard}, gradient-boosted embeddings (BIER, A-BIER) \cite{bier-pami18,opitz2017bier} are recently proposed methods that used ensemble techniques to boost the retrieval performance. 
Different from these, SGML uses a single model while achieving Recall@1 of 91.8\%, outperforming these ensemble-based methods.  
Most importantly, the proposed SGML sets a new state-of-the-art result on DeepFashion dataset outperforming the previous attention based ensemble ABE method \cite{kim2018attention} and the recent FastAP method by +4.5\% and +1\% Recall@1 respectively. This clearly demonstrates the benefit of integrating semantic similarity at multiple granularities into metric learning in SGML framework. In a nutshell, our framework is conceptually elegant, computationally simple while providing significant improvements in retrieval performance. Moreover, it is orthogonal to the ensemble-based frameworks and hence can be used to improve their performances.

\begin{table} [!t]
\footnotesize
\caption{Comparison with  state-of-the-art methods on  DeepFashion}
\label{table:df_sota}
\centering
	\begin{tabular}{l c c c c c c}
	\hline
	Recall@K 									& 1 		& 10 		& 20 		&30			&40		&50  \\ \hline 
	WTBI \cite{kiapour2015buy}  				&35.6  	&47.0 	&50.6 &52.0 &53.0 &54.4 \\
	DARN  \cite{huang2015cross} 				&38.3 	&61.0 &67.5 &70.0 &71.1 &71.8 \\
	FashionNet  \cite{liu2016deepfashion}		&53.3 	&73.0 	&76.4		& 77.0		& 79.0   	& 80.0\\ 
    
    BinDev \cite{opitz2017bier} 				& 70.6 	& 90.5 	& 93.4	&94.7 	&95.5 	&96.1 \\
    HTL \cite{ge2018deep}   					& 80.9 	& 94.3  & 95.8 	&97.2 	&97.4 	&97.8 \\
    HDC \cite{yuan2017hard}						& 62.1	& 84.9	& 89.0 	&91.2 	&92.3	&93.1 \\ 	
	BIER  \cite{opitz2017bier} 					& 76.9 	& 92.8 	& 95.2 	&96.2	&96.7	&97.1 		\\ 
	DREML \cite{xuan2018deep} 					& 78.4 	& 93.7  & 95.8 	&96.7 &- & - \\
	A-BIER \cite{bier-pami18}   				& 83.1 	& 95.1  & 96.9  &97.5 &97.8 &98.0 \\
	ABE \cite{kim2018attention}					& 87.3 	& 96.7  & 97.9 &98.2 &98.5 &98.7 \\ 
	FastAP \cite{Cakir_2019_CVPR}               &90.9   &97.7   &98.5   &98.8 &98.9 &99.1 \\ \hline
	\textbf{Proposed SGML} 					& \textbf{91.8}  & \textbf{97.7} & \textbf{98.4}  &\textbf{98.8}  &\textbf{98.9}  &\textbf{99.1}  \\
	\hline    
	\end{tabular}			
\end{table}

\subsubsection{Analysis and Ablation Studies} \hfill \par 
\label{ssec:df_analysis}
\emph{2.1) Retrieval using Features from Different Layers:}
Recent work  of  Vo and Hays \cite{emb-vo2018} showed that the last embedding feature may not be the best rather earlier layers may better generalize to the test set. Motivated by this, we conducted  retrieval experiments using features from various layers of the network.  
In particular, we performed retrieval using the embedding (\emph{Emb}), fully-connected  (\emph{FC}) features, and features pooled from the last convolutional layer (\emph{Conv}). 
Table~\ref{tab:df_layers} summarizes the retrieval results using these features. The embedding layer achieves the best Recall@1 of 91.8\% outperforming other layers by 2-3\%. Since each class in Deepfashion dataset  contains only few clothing instances with limited variations, discriminative embeddings could be learned, and hence the final \emph{Emb} layer shows better performance. 

\begin{table} [!t]
\footnotesize
\caption{Performance of the various features of the SGML network on DeepFashion.}
\label{tab:df_layers}
\centering
	\begin{tabular}{lcccccc}
	\hline
	Recall@K 			& 1 	& 10 	& 20 	&30	&40  &50  \\ \hline 
		\emph{Emb}	&\textbf{91.8}  &\textbf{97.7}  &\textbf{98.4}  &\textbf{98.8 } &\textbf{98.9}  &\textbf{99.1}  \\
		\emph{FC}	&88.0  &95.5  &96.5  &97.0  &97.3  &97.6  \\
		\emph{Conv} &89.9  &96.7  &97.5  &98.0  &98.2  &98.4  \\ \hline
	\end{tabular}
\end{table}

\emph{2.2) Impact of Sampling Method:}
\label{sssec:df_sampling}
Here we study the impact of the sampling methods used during the network training. The retrieval performances using image-wise and batch-wise sampling methods discussed in Section~\ref{sec:samplingmethods} are reported in Table~\ref{tab:df_sampling}. The image-wise achieves Recall@1 of 79.6\% whereas batch-wise sampling achieves a Recall@1 of 91.8\% which  is about 12\% better than the former.   
As DeepFashion dataset  has a large number of
classes and only few images per class, image-wise sampling
causes a severe imbalance between the putative positive and
negative pairs. On the other hand, batch-wise sampling reduces
this impact by using only a subset of classes in a batch and
considering all possible pairs within this batch. Hence, we argue that batchwise sampling has clear advantage and should be used for instance retrieval problems where only few examples are available in each class.  

\begin{table} [!t]
\footnotesize
\caption{Impact of Different Sampling Methods on  DeepFashion}
\label{tab:df_sampling}
\centering
	\begin{tabular}{l c c c c c c}
	\hline
	Recall@K 									& 1 		& 10 		& 20 		&30			&40		&50  \\ \hline
	Image-wise &79.6 &93.6 &95.4 &96.3 &96.8 &97.2 \\
	Batch-wise &\textbf{91.8}  &\textbf{97.7} &\textbf{98.4} &\textbf{98.8} &\textbf{98.9} &\textbf{99.1} \\	\hline
	\end{tabular}
\end{table}	

\begin{figure}[!t]
\centering
 \includegraphics[scale =.28]{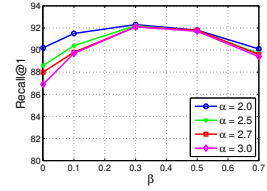}  \\
\caption{Recall@1 of the SGML for various values of $\alpha$ and $\beta$ on DeepFashion.}
\label{fig:df_paramsen}
\end{figure}

\emph{2.3) Sensitivity to $\alpha$ and $\beta$ parameters:}
The $\alpha$ and $\beta$ parameters define the steepness and translational shift of the proposed soft-binomial deviance loss curves. Here we study the performance of the SGML framework using different combinations of the parameters with $\alpha = \left\lbrace 2,2.5,2.7,3\right\rbrace$ and $\beta = \left\lbrace 0.0, 0.1, 0.3, 0.5, 0.7\right\rbrace$. Fig.~\ref{fig:df_paramsen} show  the Recall@1 results obtained using different values of $\alpha$ \& $\beta$ on DeepFashion dataset.
It is observed that $\alpha=2$ performs better than other values, however, the gap in  performances is  negligible for larger values of $\beta$. Particularly, the performance is fairly stable ($\approx$92\%) for $\beta \in [0.3,0.5]$ where the method performs the best.

\emph{2.4) Ablation Studies:}
We present step-by-step ablation studies to demonstrate the effectiveness of the proposed SGML framework. We first obtained strong baselines using a proper sampling strategy and design parameters. In particular, we compare the performances of four variants of the proposed model. (i) Metric learning network which is trained using binomial deviance loss alone. (ii) Attribute learning network that is trained with BCE loss. (iii) Multitask learning network which uses both binomial deviance loss and BCE loss where we disconnect the link between two tasks \ie do not incorporate the SGS mappings. This model reduces to a standard multitask network  that is trained using the multitask loss $L_{\text{multitask}} = L_{\text{BDL}} + \lambda L_{\text{BCE}}$. (iv) Finally, the full SGML framework using the soft-binomial deviance loss that considers semantic similarity at multiple granularities during metric learning. 

Table~\ref{table:df_sgml} summarizes the retrieval performances of the four models on DeepFashion dataset \cite{liu2016deepfashion}.  The metric learning network using the binomial deviance loss already sets a strong baseline with a Recall@1 of 89.9\% which is better than that reported in \cite{bier-pami18}. The FC layer feature from the attribute learning network  achieves a Recall@1 of 61.6\%. The multitask network
using both binomial deviance and BCE loss improves the performance to a Recall@1 of 90.5\%.  The proposed SGML learns the similarity at different granularities and further improves the performance achieving the best Recall@1 of 91.8\%. This clearly indicates the significance of the proposed SGML framework and our soft-binomial deviance loss. 

\begin{table} [!t]
\footnotesize
\caption{Ablation Studies: Comparison of baselines and the SGML framework on DeepFashion}
\label{table:df_sgml}
\centering
	\begin{tabular}{l|cccccc}
	\hline
	Recall@K 			& 1 	& 10 	& 20 	&30	&40  &50  \\ \hline
	Metric Learning 		&89.9  &97.1  &98.0  &98.4  &98.6  &98.8 \\ 
	Attribute Learning 		 &61.6  &85.5  &89.7  &91.5  &92.7  &93.5  \\ 	
	Multitask Learning 	&90.5  &97.3  &98.1  &98.4  &98.7  &98.8   \\               
	Proposed SGML	&\textbf{91.8}  &\textbf{97.7}  &\textbf{98.4}  &\textbf{98.8 } &\textbf{98.9}  &\textbf{99.1}  \\ \hline
	\end{tabular}			
\end{table}

\subsection{Results: CUB-2011-200 Dataset} 
\subsubsection{Comparison with State-of-the-art Methods}
Table~\ref{tab:cub_sota} summarizes retrieval performances on CUB dataset where the proposed SGML framework is compared with state-of-the-art methods using various sampling methods \cite{wu2017sampling,huang2016local}, loss functions \cite{opitz2017bier} \cite{huang2016local} and ensemble methods \cite{kim2018attention,bier-pami18,yuan2017hard}. 
Our SGML achieves a Recall@1 of 71.9\% outperforming methods such as distance-weighted \cite{wu2017sampling} and position-dependent sampling \cite{huang2016local}, and loss functions such as pair-wise loss \cite{wu2017sampling}, triplet/quadratic loss \cite{huang2016local} and binomial deviance loss \cite{opitz2017bier}. This clearly shows the advantage of the proposed SBDL loss and its inherent property of treating training samples based on their degree of information. Our method also outperforms the attribute-aware attention model ($A^3M$) and other ensemble based method such as hard-aware cascaded (HDC) network \cite{yuan2017hard}, gradient-boosted BIER, A-BIER \cite{opitz2017bier,bier-pami18}. Moreover, SGML outperforms the state-of-the-art attention-based ensemble (ABE) method \cite{kim2018attention} by +1.3\% Recall@1. 
This clearly demonstrates the effectiveness of the proposed SGML framework and its advantage of being computationally simple while providing  better retrieval performances.

\begin{table} [!t]
\caption{Comparison with state-of-the-art methods on cropped CUB dataset.}
\label{tab:cub_sota}
\centering
	\begin{tabular}{l c c c c c c}
	\hline
	Recall@K 									& 1 		& 2 		& 4 		&8			&16	&32  \\ \hline 
	BinDev~\cite{opitz2017bier} 				& 58.9  & 70.1 & 79.8 & 87.6  & 92.6  & 96.0\\
	 
	PDDM+Triplet \cite{huang2016local}          & 50.9  & 62.1 & 73.2 & 82.5  & 91.1  & 94.4      \\
	PDDM+Quad \cite{huang2016local}             & 58.3  & 69.2 & 79.0 & 88.4  & 93.1  & 95.7        \\
	HDC~\cite{yuan2017hard}      			    & 60.7  & 72.4 & 81.9 & 89.2  & 93.7  & 96.8      \\
	$A^3M$ \cite{han2018attribute}              &61.2   & 72.4 &81.8 &89.2 &- &- \\
	BIER~\cite{opitz2017bier} 					& 63.7  & 74.0 & 82.5 & 89.3  & 93.8  & 96.8\\
	Margin~\cite{wu2017sampling}                & 63.9  & 75.3 & 84.4 & 90.6  & 94.8  & -     \\
	A-BIER~\cite{bier-pami18}   				& 65.5  & 75.8 & 83.9 & 90.2  &94.2  &97.1 \\
	ABE~\cite{kim2018attention}		            & 70.6  & 79.8 & 86.9 &92.2   &- & -\\ \hline
	\textbf{Proposed SGML}  			&\textbf{71.9} &\textbf{81.8} &\textbf{88.5} &\textbf{93.2} &\textbf{96.2} &\textbf{98.1} \\	\hline	
	\end{tabular}			
\end{table}

\subsubsection{Analysis and Ablation Studies}  \hfill \par 
\emph{2.1) Retrieval using Features from Different Layers:}

The retrieval performances of the proposed method using features from various layers namely, \emph{Emb}, \emph{FC} and \emph{Conv} are shown in Table~\ref{tab:cub_layers}. As opposed to DeepFashion dataset, the \emph{Conv} feature achieved the best performance with a Recall@1 of 71.9\%. It is because various images from the same class in CUB dataset exhibit diversity in visual appearances especially due to the  position of the birds and backgrounds. This intra-class variance hinders in learning discriminative embeddings that can generalize all  the images in a class, potentially leading to lower performance of the embedding layer than early layers. Hence, the best performing layer of metric network depends upon the nature of the retrieval problem. Note that this observation is in-line with recent studies in \cite{emb-vo2018}. 

\begin{table} [!t]
\footnotesize
\caption{Performance of the various features of the SGML network on CUB Dataset}
\label{tab:cub_layers}
\centering
	\begin{tabular}{lcccccc}
	\hline
	Recall@K 			& 1 	& 2 	& 4 	&8	&16  &32  \\ \hline 
		\emph{Emb}	 	&64.1 &75.0 &82.8 &89.0 &93.4 &96.2  \\
		\emph{FC} 		&69.4 &78.9 &86.9 &91.9 &95.1 &97.2	 \\
		\textbf{\emph{Conv}} 	&\textbf{71.9} &\textbf{81.8} &\textbf{88.5} &\textbf{93.2} &\textbf{96.2} &\textbf{98.1}  \\ \hline
	\end{tabular}
\end{table}

\emph{2.2) Impact of Sampling Methods:}
The performance of the SGML framework using the image-wise and batch-wise sampling methods is presented in Table~\ref{tab:cub_sampling}. Since CUB dataset contains relatively large number of images per class, sufficient positive pairs can be sampled using both image-wise and batch-wise sampling method \ie there is no severe imbalance between possible candidates for positive and negative pairs. Hence, both sampling methods achieves similar retrieval performance with a Recall@1 of 71\%. 

\begin{table} [!t]
\footnotesize
\caption{Impact of different sampling methods on CUB dataset.}
\label{tab:cub_sampling}
\centering
	\begin{tabular}{l c c c c c c}
	\hline
	Recall@K 	& 1  & 2 	& 4 		&8		&16		&32  \\ \hline
	Image-wise   &\textbf{71.9} &\textbf{81.8} &\textbf{88.5} &\textbf{93.2} &\textbf{96.2} &\textbf{98.1} \\
	Batch-wise  &71.2 &80.8 &88.2 &93.1 &95.9 &97.8 \\	\hline
	\end{tabular}
\end{table}	

\begin{figure}[!t]
\centering
 \includegraphics[scale =.28]{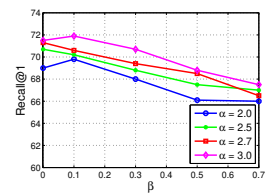}  \\
\caption{Recall@1 of the SGML  for various values of $\alpha$ and $\beta$ on CUB dataset.}
\label{fig:cub_paramsen}
\end{figure}

\emph{2.3) Sensitivity to $\alpha$ and $\beta$ Parameters:}
Fig.~\ref{fig:cub_paramsen} shows Recall@1 obtained for various values of $\alpha$ and $\beta$ on CUB dataset. Only marginal variations in performance can be observed with different $\alpha$ values. Using $\alpha = 3$ shows better performance, and achieves the peak Recall@1 of 71.9\% at $\beta = 0.1$.

\emph{2.4) Ablation Studies:}
Here we present ablation studies comparing various models similar to Section~\ref{ssec:df_analysis}.4. Table~\ref{table:cub_sgml} summarizes the retrieval performances on CUB dataset.
The metric learning network using binomial deviance loss alone achieves a strong baseline of Recall@1 of 66.6\% which is already better than the baseline (58.9) reported in  \cite{bier-pami18}. 
The attribute learning network  achieves a competitive performance with Recall@1 of 64.6\%. The multitask network further improves the performance to 67.7\%.
Most importantly, the proposed SGML framework achieves a Recall@1 of 71.9\% which is about +5\% and +4\% improvement over the  binomial deviance loss and the multitask learning respectively. This clearly shows the effectiveness of the SGML framework.

\begin{table} [!t]
\footnotesize
\caption{Ablation Studies: Comparison of baselines and the SGML framework on CUB dataset.}
\label{table:cub_sgml}
\centering
	\begin{tabular}{l|cccccc}
	\hline
	Recall@K 			& 1 	& 2 	& 4 	&8	&16  &32   \\ \hline
	Metric Learning 	&66.6 &77.5 &86.0 &91.5 &95.5 &97.5	\\ 
	Attribute Learning 	&64.6 &75.1 &83.9 &90.0 &94.1 &96.8	  \\ 	
	Multitask Learning 	&67.7 &78.3 &86.6 &92.1 &95.6 &97.7  \\               
	Proposed SGML	&\textbf{71.9} &\textbf{81.8} &\textbf{88.5} &\textbf{93.2} &\textbf{96.2} &\textbf{98.1}  \\ \hline
	\end{tabular}			
\end{table}

\begin{figure}
\centering
\begin{tabular}{c}
\includegraphics[scale =0.32]{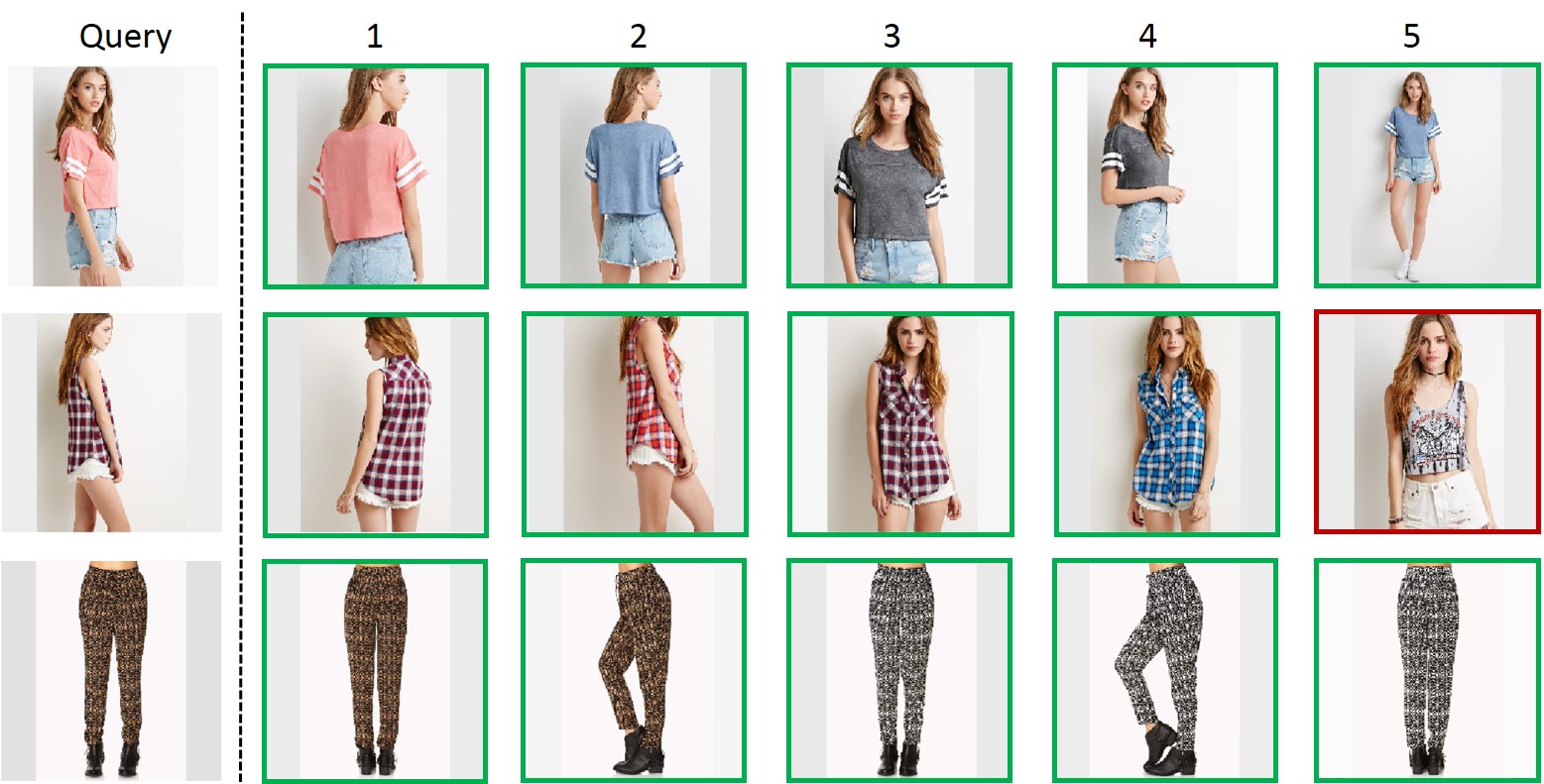} \\
(a) DeepFashion Inshop dataset \cite{liu2016deepfashion}  \\
\includegraphics[scale=0.32]{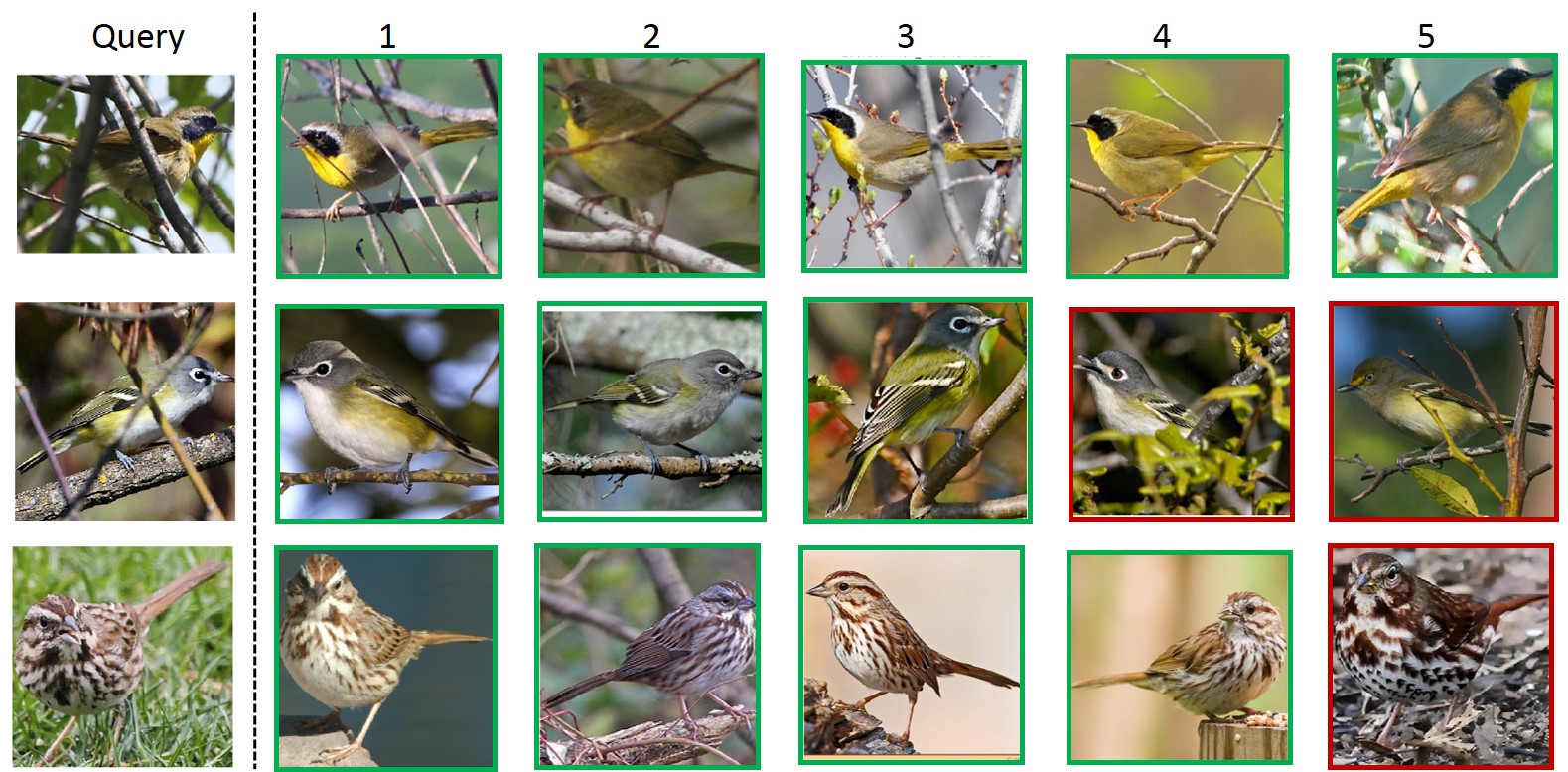}\\
(b) CUB-200-2011  dataset \cite{WelinderEtal2010}
\end{tabular}
\caption{Qualitative retrieval results for (a) DeepFashion and (b) CUB dataset. 
Top 5 most visually similar images to the query are shown; correct results are highlighted in green and incorrect results are highlighted in red.}
\label{fig:retrievals}
\end{figure}

Overall, our SGML framework performs very well, outperforming  other  existing methods and establishing new state-of-the-art results on standard benchmark datasets. The experimental results clearly demonstrates the effectiveness of the proposed SGML framework. Fig.~\ref{fig:retrievals} shows qualitative retrieval results for the proposed SGML method on DeepFashion dataset and CUB dataset. 
For both datasets, our method successfully retrieves images from the same query class/ ID  on the top ranks. For example, in the first row of Fig.~\ref{fig:retrievals}(a), the exact instance of \emph{pink-top} is retrieved first followed by other \emph{tops} of the same clothing ID. This further illustrates the effectiveness of the proposed method.

\section{Conclusion}
\label{sec:conclusion} 
This paper proposed a new semantic granularity metric learning for visual search. 
We introduced a novel idea of detecting and integrating  attribute semantic space information into metric learning that helps to capture the underlying similarity at different granularities found in many domain applications. 
The SGML framework  interlinks multitask and metric learning using the proposed semantic attribute similarity mappings.
We  proposed a new soft-binomial deviance metric loss function which is dynamically learned based on  the degree of information in  training samples.
Overall, the proposed SGML framework is conceptually elegant, computationally simple and provides significant improvements in retrieval performances. 
Experiments  on the benchmark  datasets show that our method outperforms the existing methods and achieves state-of-the-art results which clearly demonstrate the effectiveness of the proposed SGML framework.

\appendices
\ifCLASSOPTIONcaptionsoff
  \newpage
\fi

\bibliographystyle{IEEEtran}
\bibliography{SGML_dipu}

\end{document}